\pdfoutput=1
\PassOptionsToPackage{svgnames}{xcolor}
\documentclass[11pt]{article}

\usepackage[final]{acl}

\usepackage{times}
\usepackage{latexsym}

\usepackage[T1]{fontenc}
\usepackage[utf8]{inputenc}

\usepackage{microtype}

\usepackage{inconsolata}
\usepackage{booktabs}
\usepackage{multirow}
\usepackage{amsmath}
\usepackage{amsfonts}
\usepackage{amsthm}
\usepackage{pifont}
\usepackage{xspace}
\usepackage{graphicx}
\usepackage{bm}
\usepackage{colortbl}
\usepackage[most]{tcolorbox}
\usepackage{titlesec}
\usepackage{csquotes}
\usepackage{anyfontsize}
\usepackage{comment}
\usepackage{float}
\usepackage{cuted}
\usepackage{tikz}
\usepackage{caption}
\usepackage{subcaption}
\usepackage{enumitem}
\usepackage{cleveref}
\usepackage{stmaryrd}
\usepackage{hhline}

\usepackage{color-edits}

\addauthor{gn}{magenta}

\newtcolorbox[list inside=prompt,auto counter]{prompt}[1][]{
    colbacktitle=black!60,
    coltitle=white,
    fontupper=\footnotesize,
    boxsep=5pt,
    left=0pt,
    right=0pt,
    top=0pt,
    bottom=0pt,
    boxrule=1pt,
    #1,
}

\newcommand{\dataset}[1]{\texttt{#1}}

\newcommand{\numone}{\xspace\ding{192}\xspace}
\newcommand{\numtwo}{\xspace\ding{193}\xspace}
\newcommand{\numthree}{\xspace\ding{194}\xspace}
\newcommand{\numfour}{\xspace\ding{195}\xspace}
\newcommand{\numfive}{\xspace\ding{196}\xspace}
\newcommand{\numsix}{\xspace\ding{197}\xspace}
\newcommand{\numseven}{\xspace\ding{198}\xspace}
\newcommand{\numeight}{\xspace\ding{199}\xspace}

\newcommand{\hlcell}{\cellcolor{AliceBlue!50}}
\newcommand{\hlcelll}{\cellcolor{LavenderBlush!50}}
\newcommand{\hlcellll}{\cellcolor{LightGreen!20}}

\def\mediumcol{\hskip 6pt}
\def\smallcol{\hskip 4pt}

\newcommand{\nextphase}{$\shortrightarrow$\xspace}

\newcommand{\ft}{instruction-tuning\xspace}
\newcommand{\preft}{pre-instruction-tuning\xspace}
\newcommand*{\pit}{PIT\xspace}

\titlespacing*{\section}{0pt}{0.5ex plus 0.3ex minus 0.3ex}{0.5ex plus 0.1ex minus 0.1ex}
\titlespacing*{\subsection}{0pt}{0.4ex plus 0.4ex minus 0.2ex}{0.2ex plus 0.1ex minus 0.1ex}
\title{Instruction-tuned Language Models are Better Knowledge Learners}

\author{Zhengbao Jiang$^{2}$\thanks{Majority of the work done during an internship at Meta.} \quad Zhiqing Sun$^2$ \quad Weijia Shi$^{1,3}$ \quad Pedro Rodriguez$^1$ \quad Chunting Zhou$^1$ \\
{ \bf Graham Neubig$^2$ \quad Xi Victoria Lin$^1$ \quad Wen-tau Yih$^1$  \quad Srinivasan Iyer$^1$} \\
$^1$FAIR at Meta \quad $^2$Carnegie Mellon University \quad $^3$University of Washington \\
\texttt{\{zhengbaj,gneubig\}@cs.cmu.edu} \quad \texttt{\{victorialin,scottyih,sviyer\}@meta.com}}

\begin{document}
\maketitle
\begin{abstract}
In order for large language model (LLM)-based assistants to effectively adapt to evolving information needs, it must be possible to update their factual knowledge through continued training on new data.
The standard recipe for doing so involves continued pre-training on new documents followed by \ft on question-answer (QA) pairs.
However, we find that LLMs trained with this recipe struggle to answer questions, even though the perplexity of documents is minimized.
We found that QA pairs are generally straightforward, while documents are more complex, weaving many factual statements together in an intricate manner.
Therefore, we hypothesize that it is beneficial to expose LLMs to QA pairs \emph{before} continued pre-training on documents so that the process of encoding knowledge from complex documents takes into account how this knowledge is accessed through questions.
Based on this, we propose \textbf{\preft (\pit)}, a method that instruction-tunes on questions prior to training on documents.
This contrasts with standard \ft, which learns how to extract knowledge after training on documents.
Extensive experiments and ablation studies demonstrate that \pit significantly enhances the ability of LLMs to absorb knowledge from new documents, outperforming standard \ft by 17.8\%.
%\footnote{Code and datasets are available at \url{https://anonymous}.}
\end{abstract}

\begin{figure*}[tb]
\includegraphics[width=0.8\textwidth, clip, keepaspectratio]{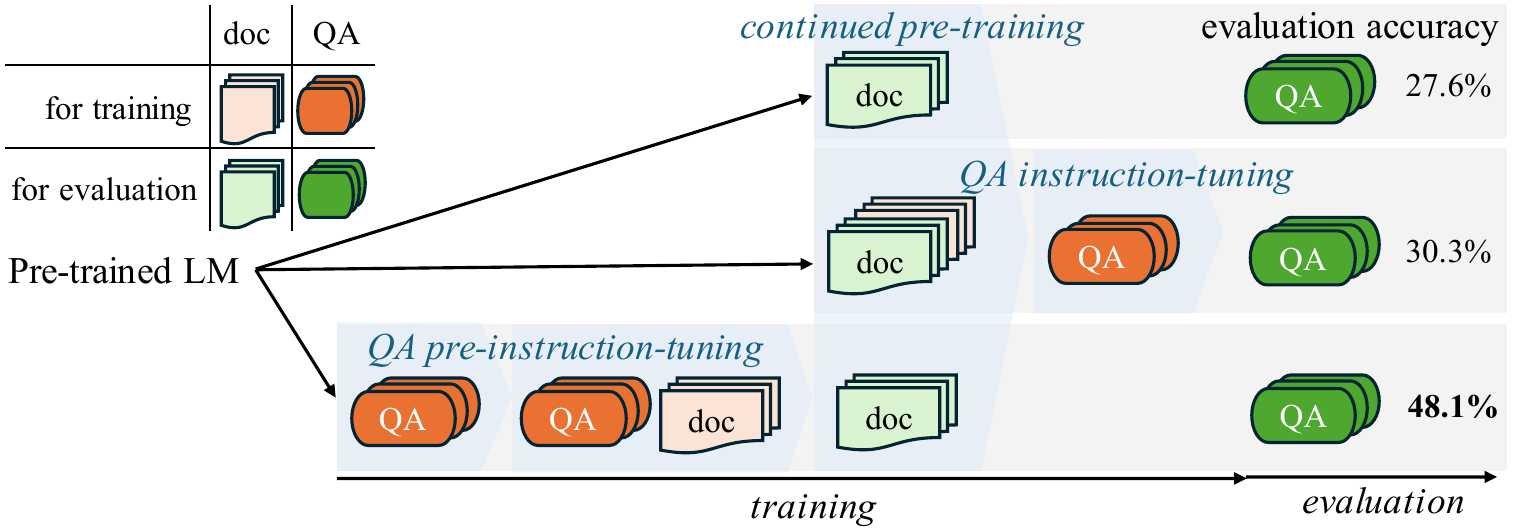}
\centering
\caption{Illustration of continued pre-training (first row), continued pre-training followed by \ft (second row), and \preft before continued pre-training (last row), along with their accuracies on evaluation questions. Each right-pointing light-blue triangle indicates a training phase.}
\label{fig:thumbnail}
\end{figure*}

\section{Introduction}
Large language models (LLMs) store vast amounts of factual knowledge in their parameters through large-scale pre-training, and this knowledge can be used to answer various questions such as ``where is the world's largest ice sheet located'' \cite{gpt3-brown-2020,gpt4-2023,palm-chowdhery-2022,opt-zhang-2022,llama-touvron-2023,llama2-touvron-2023,gemini-2023}.
However, this factual knowledge is static, meaning that it can become outdated as the world evolves, or prove insufficient when LLMs are used in specialized or private domains.

To keep LLMs up-to-date, it is common to continue pre-training on new documents to store knowledge in parameters, which allows LLMs to effectively answer queries that require up-to-date information \cite{continual-jang-2023}.
A widely held view is that the factual knowledge stored in parameters can be elicited through prompting \cite{gpt3-brown-2020,lama-petroni-2019,t5pack-roberts-2020}, and that \ft (also known as supervised fine-tuning or alignment) makes this elicitation more effective \cite{t0-sanh-2022,flan-jason-2022,instructgpt3-ouyang-2022}.
In the first part of this paper (\autoref{sec:standard}), we conduct extensive experiments using Llama-2 \cite{llama2-touvron-2023} to answer the following question: \emph{to what extent can we augment the knowledge stored in modern LLMs by continued pre-training on new documents, either with or without subsequent \ft}?
We find that, as we train LLMs repeatedly over documents to the extent that perplexity is minimized to one, the percentage of questions regarding those documents that LLMs answer correctly increases consistently to 27.6\%.
Subsequent \ft further improves it to 30.3\%, confirming that this widely used practice is useful to elicit more knowledge from LLMs.\footnote{This capacity might be underestimated by previous works due to using relatively small LMs or randomly initialized transformers, or lack of exhaustive training or instruction-tuning \cite{lmkb-wang-2021,camels-hu-2023,part1-zhu-2023}.}
However, the amount of elicited knowledge is still limited, even though the perplexity of documents is minimized, a phenomenon we refer to as the ``perplexity curse''.\footnote{Inspired by the ``reversal curse'' of \citet{reversal-berglund-2023}.}

In the second part of the paper (\autoref{sec:pre}), we study methods to mitigate the perplexity curse by making LLMs more adept at absorbing knowledge from documents.
\citet{part1-zhu-2023} presented an intriguing finding that training a randomly initialized transformer from scratch on a mix of biographies and related questions resulted in strong generalization to new questions.
However, understanding the reasons behind this finding and exploring ways to practically apply it for absorbing knowledge from new documents requires further investigation.
We found that question-answer (QA) pairs are generally straightforward and easily digestible, while documents tend to be more complex and cluttered, often weaving many factual statements together in a more intricate manner.
Therefore, we hypothesize that \emph{it is beneficial to deliberately expose LLMs to QA data before continued pre-training on documents so that the process of encoding knowledge from complex documents takes into account how this knowledge is accessed through questions}.
We refer to this as \textbf{\preft (\pit)} and conduct comprehensive experiments to benchmark different variations of this method.
As shown in \autoref{fig:thumbnail}, our best-performing variation starts with training exclusively on QA pairs (e.g., ``who handled the editing of Oppenheimer'') to grasp how knowledge is accessed.
This is followed by training on a combination of these QA pairs and associated documents (e.g., ``who handled the editing of Oppenheimer'' and a document about ``Oppenheimer'').
In this phase, LLMs enhance their ability to absorb knowledge from information-dense documents, building upon the QA pairs that they have already mastered.
To study continual knowledge acquisition, we build a dataset named \dataset{Wiki2023}, which includes a collection of documents from Wikipedia that are relevant to the year 2023.
Comprehensive experiments on \dataset{Wiki2023} demonstrate that after \pit, LLMs exhibit an enhanced ability to absorb knowledge from new documents (e.g., a document about ``Barbie'').
Detailed ablation studies reveal that this ability primarily stems from prioritizing learning how to access knowledge over learning to encode knowledge from documents.
Overall, \pit significantly outperforms the standard \ft approach (\autoref{sec:pre_method} and \autoref{sec:pre_plus}), improving QA accuracies by 17.8\% on Llama-2 7B (30.3\% \nextphase 48.1\%) and 16.3\% on Llama-2 70B (46.4\% \nextphase 62.7\%).
Moreover, \pit also enhances the ability to absorb knowledge from documents of a \emph{different} domain, shedding light on the potential to scale this method up to a wider variety of documents and instructions for more robust generalization (\autoref{sec:pre_ood}).

\section{Building a Dataset to Study Continual Knowledge Acquisition}\label{sec:dataset}
To assess the ability of LLMs to learn knowledge from new documents, it is essential to use a document corpus with minimal overlap with the original pre-training corpus.
This ensures that when an LLM correctly answers questions, we can confidently attribute this capability to its learning from the new documents, rather than encountering similar questions in its original pre-training corpus.
In this section, we describe a methodology for building such a corpus from Wikipedia.

\begin{figure}[!tb]
\centering
\begin{subfigure}[t]{1.0\columnwidth}
\centering
\includegraphics[width=1.0\linewidth]{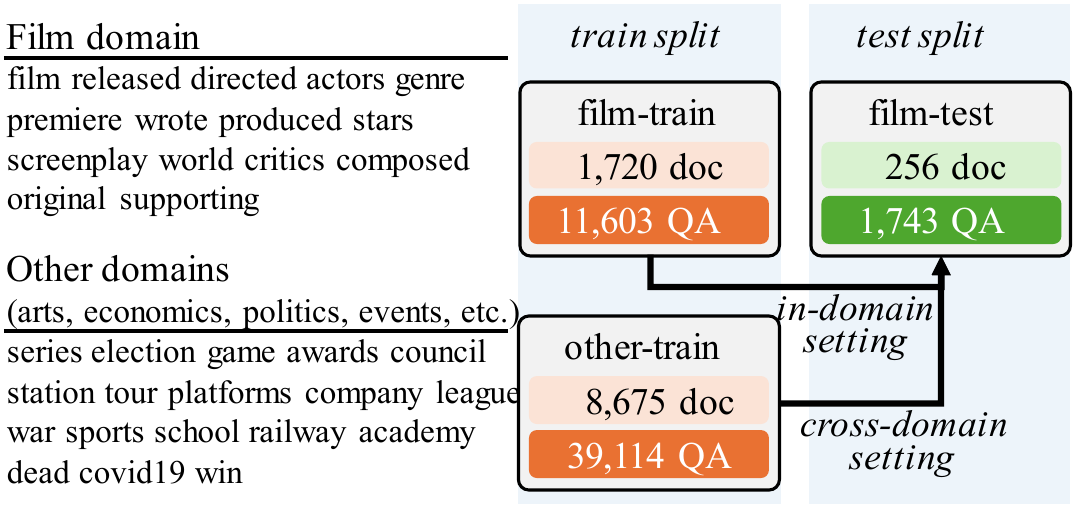}
\end{subfigure}
\begin{subfigure}[t]{1.0\columnwidth}
\centering
\includegraphics[width=1.0\linewidth]{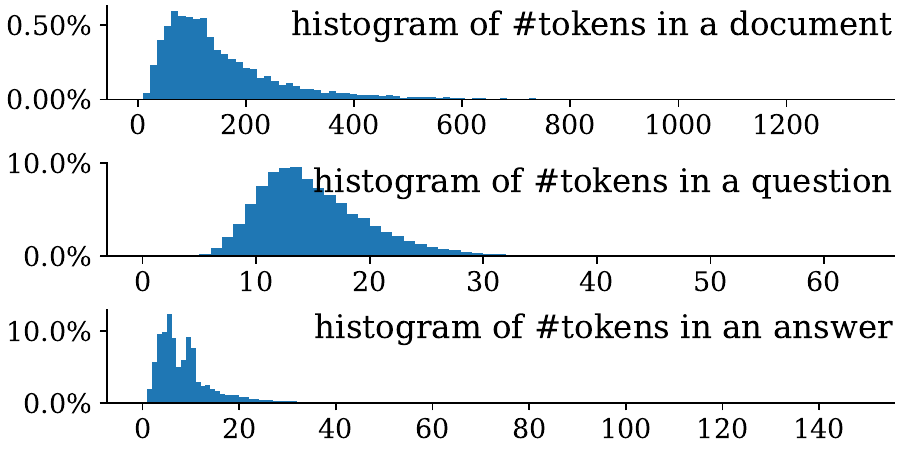}
\end{subfigure}
\caption{The \dataset{Wiki2023} dataset. \textbf{Top-right}: the number of documents and QA pairs; \textbf{Top-left}: frequent keywords in questions; \textbf{Bottom}: the distribution of token counts in documents, questions, and answers.}
\label{fig:dataset}
\end{figure}

\begin{figure}[tb]
\includegraphics[width=1.0\columnwidth, clip, keepaspectratio]{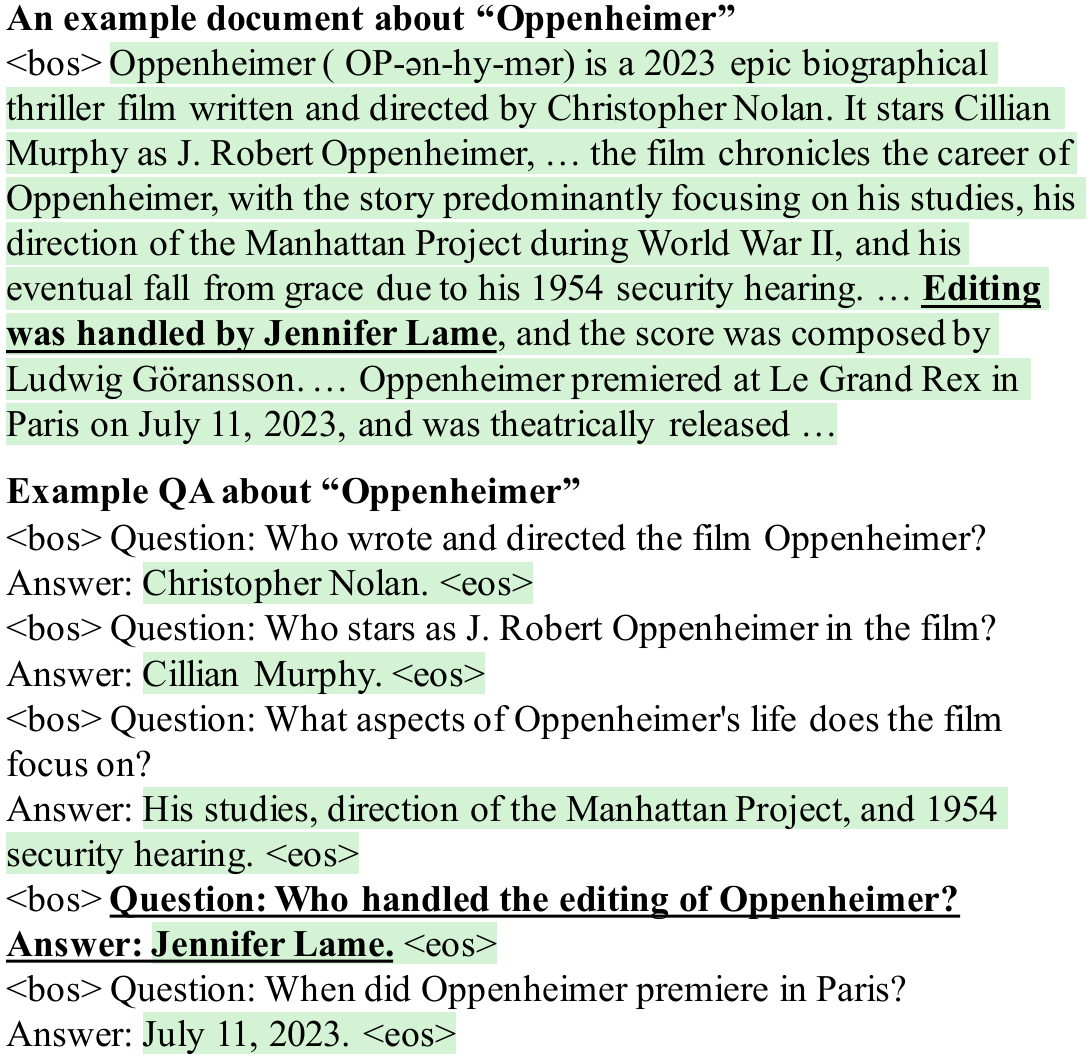}
\centering
\caption{An example document about ``Oppenheimer'' and corresponding QA pairs from \dataset{Wiki2023}. Tokens used for computing losses are highlighted in green.}
\label{fig:example}
\end{figure}

\subsection{Wiki2023 Document Corpus}
In the following experiments (\autoref{sec:standard} and \autoref{sec:pre}), we use Llama-2 (7B and 70B) \cite{llama2-touvron-2023} since it is one of the best-performing LLMs.
We use Wikipedia articles classified under the ``2023'' Category including topics from diverse domains such as films, arts, economics, politics, events, etc.\footnote{\url{https://en.wikipedia.org/wiki/Category:2023}}
The likelihood that this factual information is not included in the original training corpus is supported by the low QA performance in \autoref{tab:standard} (9.5\%/17.2\% for 7B/70B).%
\footnote{It is important to note the difficulty in completely avoiding factual overlap between \dataset{Wiki2023} and the pre-training corpus of Llama-2.
For example, a film released in 2023 might have had information available before 2023.
Data duplication detection is an active research direction, which falls beyond the focus of this study.}
To accelerate the training process, we only use the first section of each article, which offers a thorough summary and contains many factual statements.
The number of collected documents and an example document about ``Oppenheimer'' can be found in \autoref{fig:dataset} and \autoref{fig:example}.
We refer to this as the \dataset{Wiki2023} dataset.

\subsection{Wiki2023 Question-answer Pairs}
To collect QA pairs for either \ft or performance evaluation, we employ publicly available LLMs to generate diverse questions and their respective answers given the article as context, following the Prompt~\autoref{prompt:qa} in \autoref{sec:app_wiki2023}.
On average, 4.93 questions are generated for each article. \autoref{fig:dataset} and \autoref{fig:example} show the detailed statistics and example QA pairs about ``Oppenheimer'', respectively.

\subsection{Splits}
Among all domains, we select the film domain for evaluation and randomly select 256 articles as the test split (\dataset{Wiki2023-film-test}).
We continually train LLMs on documents from the test split (\dataset{Wiki2023-film-test-doc}), and assess their performance based on the accuracy of corresponding questions (\dataset{Wiki2023-film-test-QA}).
The remaining 1720 articles and corresponding QA pairs (\dataset{Wiki2023-film-train}) will be used to study different training strategies, which corresponds to the in-domain setting in \autoref{fig:dataset}.
We also train on other domains before evaluation on the film domain to study the effectiveness of different methods across domains, which corresponds to the cross-domain setting in \autoref{fig:dataset}.

\begin{figure*}[tb]
\includegraphics[width=1.0\textwidth, clip, keepaspectratio]{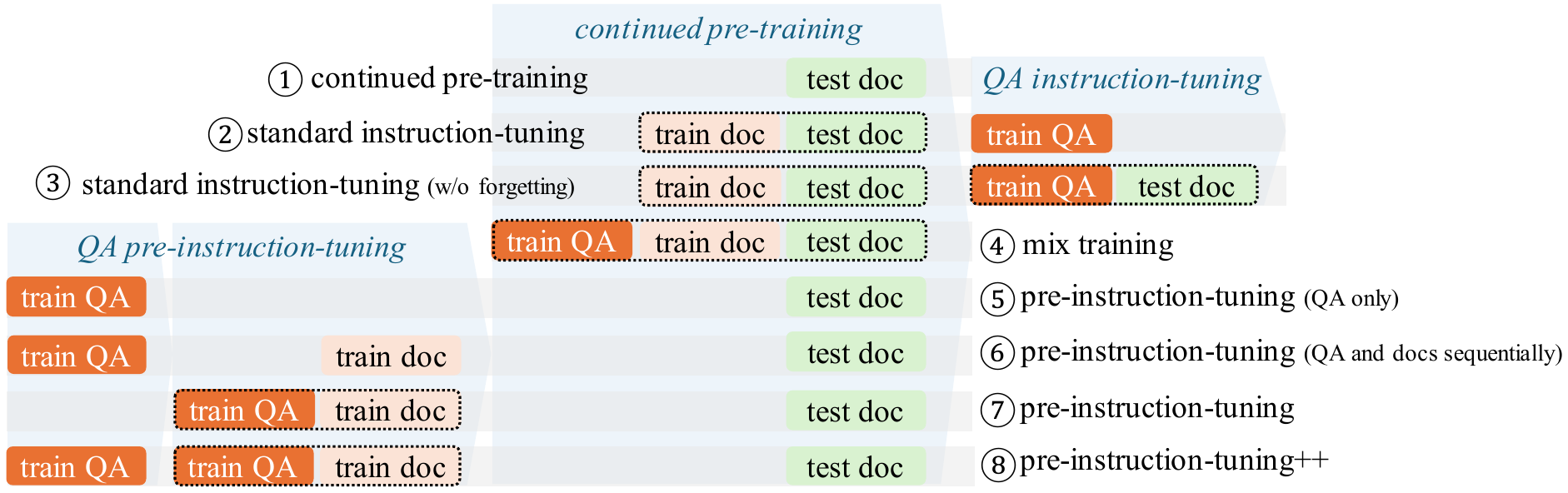}
\centering
\caption{Different experimental settings examined in this paper. Each row represents a different experimental setting with a unique name and number, and each vertical section highlighted by a right-pointing light-blue triangle indicates a training phase. Models are assessed on test QA across all settings. Whenever multiple datasets are enclosed within a dashed square, they are mixed together during the training process.}
\label{fig:settings}
\end{figure*}

\section{Experimental Settings}
\subsection{Objectives}
When training on documents, we prepend a <bos> token and compute the standard next-token prediction loss by averaging over all tokens in the document: $L_{\bm{d}} = -\sum_{t}{\log P(\bm{d}_t|\bm{d}_{<t})} / |\bm{d}|$.\footnote{We do not append a <eos> token at the end of documents because we only use the first section, which does not signify the conclusion of the entire article.}
When training on QA pairs, we compute the average negative log-likelihood loss only on tokens in the answer given the question as the prefix: $L_{\bm{a}} = -\sum_{t}{\log P(\bm{a}_t|\bm{q}, \bm{a}_{<t})} / |\bm{a}|$.
\autoref{fig:example} presents an example document alongside QA pairs, where tokens used for computing losses are highlighted.

\subsection{Hyperparameters}\label{sec:hyper}
When pre-training on documents, we use a batch size of 256 documents and an initial learning rate of 3e-5.
During \ft on QA pairs, we use the same batch size of 256 QA pairs but opt for a reduced initial learning rate of 5e-6 because the number of tokens in a single batch is lower.
Details can be found in \autoref{sec:app_hyper}.

\subsection{Evaluation Metrics}
Since most answers are relatively short, we use exact match (EM) as our primary metric \cite{nq-kwiatkowski-2019}.
To assess longer responses and accommodate minor lexical differences, we also report answer recall and ROUGE-L.
Details can be found in \autoref{sec:app_metric}.

\begin{figure*}[tb]
\centering
\begin{subfigure}[t]{0.48\textwidth}
\centering
\includegraphics[width=1.0\linewidth]{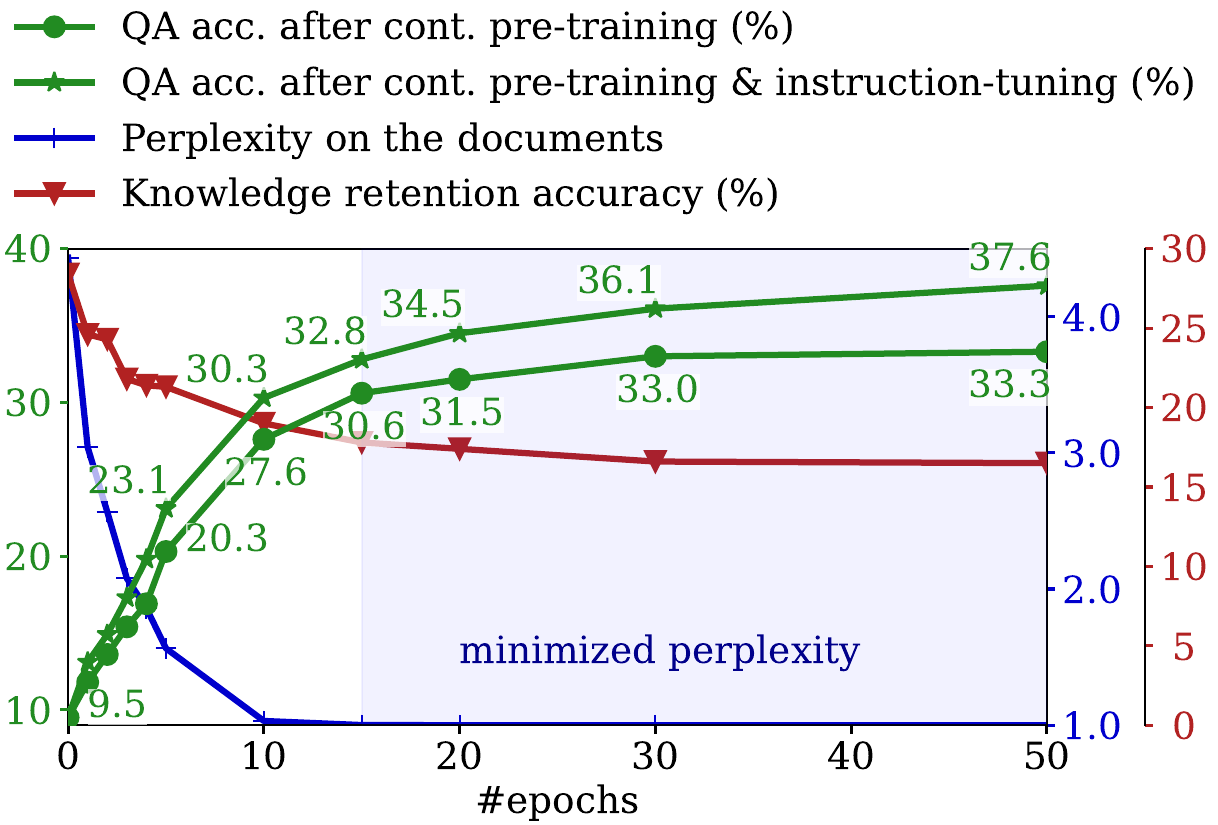}
\caption{Training dynamics w/ (\autoref{fig:settings} \numtwo) and w/o \ft (\autoref{fig:settings} \numone). Reduction in perplexity consistently leads to improvement in QA accuracy, indicating that factual knowledge acquisition necessitates exhaustive loss minimization.}
\label{fig:epochs}
\end{subfigure}\hfill%
\begin{subfigure}[t]{0.50\textwidth}
\centering
\includegraphics[width=1.0\linewidth]{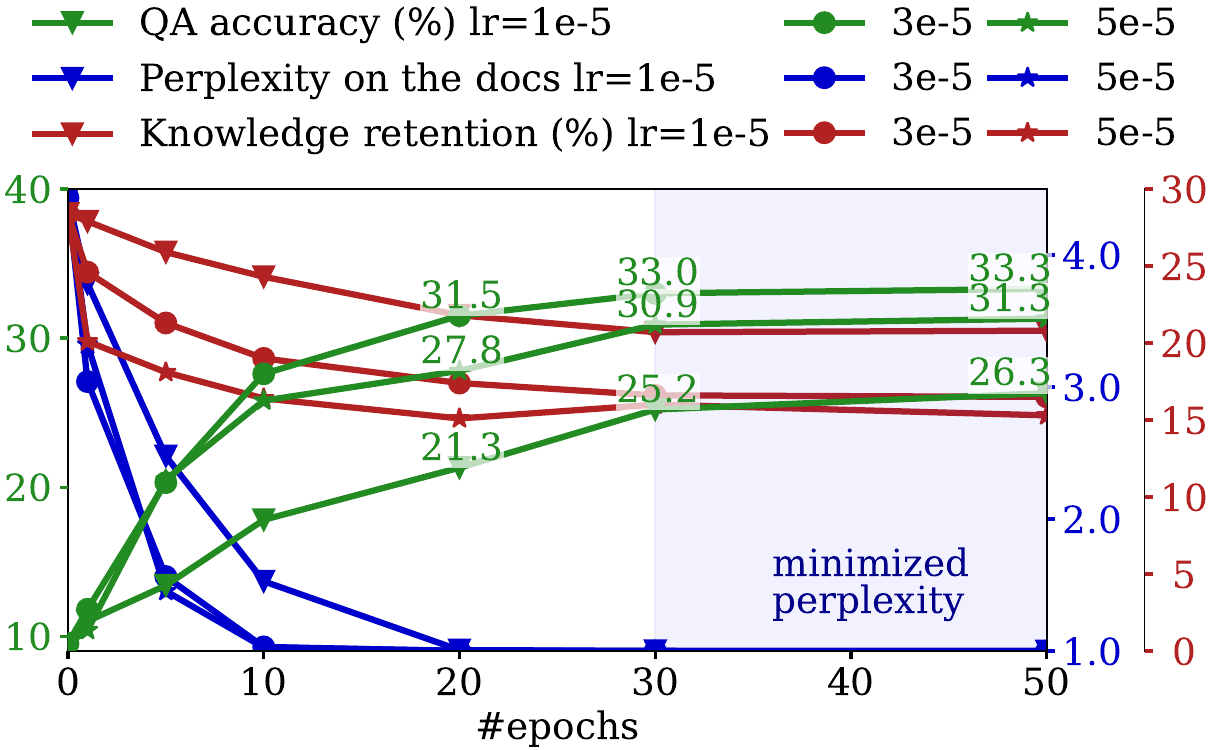}
\caption{Training dynamics with different learning rates (\autoref{fig:settings} \numone). After perplexity is minimized, larger learning rates usually lead to less overfitting to deceptive patterns in documents and better generalization when responding to questions.}
\label{fig:lr}
\end{subfigure}
\caption{We vary the number of epochs (\autoref{fig:epochs}) and learning rate (\autoref{fig:lr}) during continued pre-training to study the training dynamics of Llama-2 7B. The left axis is \textcolor{ForestGreen}{QA accuracies} for test questions, measured by exact match. On the right axis, we display 2 metrics indicated by distinct colors: the \textcolor{MediumBlue}{perplexity} of all tokens in the documents, and the \textcolor{FireBrick}{knowledge retention accuracy}, measured by QA accuracy on the Natural Questions dataset. We highlight situations where \colorbox{blue!5}{\textcolor{DarkBlue}{perplexity of all document tokens is minimized to 1}}.}
\label{fig:dynamic}
\end{figure*}

\section{How Much Knowledge Can LLMs Absorb via Continued Pre-training Followed by \expandafter\MakeUppercase\ft?}\label{sec:standard}
Factual knowledge stored in the parameters of LLMs can be accessed and applied to answering questions through prompting without additional training \cite{gpt3-brown-2020,lama-petroni-2019,lpaqa-jiang-2020,t5pack-roberts-2020}.
With additional \ft (also known as supervised fine-tuning) on high-quality data \cite{t0-sanh-2022,flan-jason-2022}, knowledge seems to be more effectively elicited from LLMs.
However, when LLMs correctly answer a question, the source of the knowledge is unclear due to the diversity of the pre-training data.
For instance, when answering the question ``where is the world's largest ice sheet located'', do LLMs derive their response by recalling and generalizing information from a seen document about the Antarctic ice sheet, or do they merely repeat answers from similar questions encountered in the training data?
This distinction is crucial, as the former scenario implies an ability to comprehend documents and effectively store knowledge within parameters in a way that can be elicited later, whereas the latter is mere rote memorization.

Several works have studied this problem and the predominant finding is that LMs struggle to answer questions about documents they have been trained on \cite{lmkb-wang-2021,part1-zhu-2023}.
It is important to note, however, that these experiments were mainly conducted using relatively small LMs such as BART, T5, or GPT-2 \cite{lmkb-wang-2021,continual-jang-2023,camels-hu-2023}, using randomly initialized transformers \cite{part1-zhu-2023}, or without \ft \cite{ftrag-ovadia-2023}.
This makes us wonder \emph{what are the actual limits of modern LLMs to absorb knowledge from new documents and answer questions about them using the standard continued pre-training followed by \ft recipe}.
In this section, we run extensive experiments using Llama-2 7B and 70B on \dataset{Wiki2023-film} to test their limits.

\subsection{Vanilla Continued Pre-training and \expandafter\MakeUppercase\ft}
\paragraph{Experimental settings} We experiment with two standard settings and assess their performance by answering associated questions.
\begin{itemize}[leftmargin=*]
\item Continued pre-training: train on test documents without \ft (\autoref{fig:settings} \numone).\footnote{We found that LLMs struggle to adhere to the QA format after training on raw documents for multiple epochs. Therefore, we include a small set of QA pairs (64) during continued pre-training to prevent LLMs from forgetting the QA format.}
\item Standard \ft: train on both train and test documents before \ft on train QA pairs (\autoref{fig:settings} \numtwo).
\end{itemize}
We perform \ft for a single epoch since more epochs usually result in diminished performance.
For training on documents, we opt for multiple epochs (10/5 for a 7B/70B model), which allows for effective knowledge acquisition and remains affordable for corpora of moderate sizes.

\paragraph{Experimental results} As shown in \autoref{tab:standard}, the relatively low performance of the original Llama-2 model (9.5\%/17.2\% for 7B/70B) indicates that most knowledge in the test documents is not included in the original pre-training corpus.
After continued pre-training on documents, performances increase to 27.2\%/41.7\%, indicating that LLMs can absorb some amount of knowledge.
\expandafter\MakeUppercase\ft further increases the performance to 30.3\%/46.4\%, confirming the effectiveness of this standard recipe.
This observation is different from \citet{part1-zhu-2023}, which demonstrates that \ft after pre-training is ineffective on a randomly initialized GPT-2-like transformer.
The difference probably arises because Llama-2, through its pre-training on diverse corpora comprising raw documents and QA data, has developed a certain degree of proficiency in extracting knowledge from its parameters via questions.
We also report the performance where the corresponding document is directly provided to Llama-2 as context (``open-book w/ doc'' in \autoref{tab:standard}).
The significant gap between closed-book and open-book settings suggests that retrieving knowledge from the parameters of LLMs is still challenging.

\subsection{Analyzing the Training Dynamics: Perplexity and Generalization}
How does lower perplexity of documents lead to generalization to answering related questions?
We vary the number of epochs (\autoref{fig:epochs}) and learning rate (\autoref{fig:lr}) for continued pre-training on documents and monitor three metrics to study the training dynamics.\footnote{Since we always decay the learning rate to 10\% of its initial value, training for more epochs is not the same as continuing training from a checkpoint obtained after fewer epochs.}
\begin{itemize}[leftmargin=*]
\item \textbf{Knowledge acquisition} QA accuracies on test questions measured by exact match.
\item \textbf{Perplexity of documents} We compute perplexity (PPL) on all tokens within the documents.
\item \textbf{Knowledge retention} We approximate the retention of accumulated knowledge during pre-training by assessing the QA accuracy on the Natural Questions (NQ) dataset. NQ was released in 2019, and primarily includes questions based on Wikipedia articles from that time.
\end{itemize}

\begin{table}[tb]
\small
\centering
\begin{tabular}{@{}l@{\smallcol}r@{\smallcol}r@{\smallcol}r|r@{\smallcol}r@{\smallcol}r@{}}
\toprule
 & \multicolumn{3}{c|}{\textbf{Llama-2 7B}} & \multicolumn{3}{c}{\textbf{Llama-2 70B}} \\
\textbf{Settings} & \textbf{EM} & \textbf{Rec.} & \textbf{R-L} & \textbf{EM} & \textbf{Rec.} & \textbf{R-L} \\
\midrule
\multicolumn{7}{c}{\emph{closed- and open-book performance before training}} \\
closed-book & 9.5 & 10.0 & 21.2 & 17.2 & 18.1 & 31.4 \\
open-book w/ doc & 72.2 & 75.4 & 91.5 & 78.2 & 80.6 & 94.9 \\
\midrule
\multicolumn{7}{c}{\emph{closed-book performance w/ standard methods}} \\
cont. pre-training \numone & 27.6 & 31.6 & 43.8 & 41.7 & 45.8 & 60.2 \\
\quad+\ft \numtwo & 30.3 & 34.7 & 47.4 & 46.4 & 50.9 & 64.1 \\
mix all data \numfour & 39.4 & 44.6 & 56.7 & 57.1 & 63.4 & 72.4 \\
\midrule
\multicolumn{7}{c}{\emph{closed-book performance w/ \preft (\pit)}} \\
\pit\space{\small (QA only)} \numfive & 28.6 & 32.7 & 45.2 & 49.7 & 53.7 & 67.9\\
\pit\space{\small (QA \nextphase docs)} \numsix & 32.5 & 37.2 & 49.0 & 54.6 & 60.0 & 73.8\\
\pit \numseven & \textbf{45.4} & \textbf{51.2} & \textbf{63.2} & \textbf{62.7} & \textbf{68.6} & \textbf{78.8} \\
\bottomrule
\end{tabular}
\caption{Comparison of QA performance (\%) between standard \ft and \preft. The best results are in bold. Rec. is short for answer recall, and R-L refers to ROUGE-L.}
\label{tab:standard}
\end{table}

\paragraph{Experiment results}
\begin{itemize}[leftmargin=*]
\item As shown in \autoref{fig:epochs}, QA accuracy consistently improves as perplexity approaches one, indicating that \emph{factual knowledge learning necessitates exhaustive loss minimization over all tokens}.
This contrasts with learning general skills, where overly optimizing leads to overfitting.
\item As shown in \autoref{fig:epochs} and \autoref{fig:lr}, among all cases where LLMs have minimized perplexity on documents, for reasonably small learning rates (5e-5 is too large and leads to overfitting), cases trained with more epochs or larger learning rates typically exhibit superior QA performance. We hypothesize that \emph{more aggressive training leads to less overfitting to deceptive patterns in documents and better generalization when responding to questions}.
\end{itemize}
In summary, lower perplexity does lead to stronger generalization when responding to questions, but it comes at the expense of forgetting previously acquired knowledge.

\begin{comment}
\begin{table}[tb]
\small
\centering
\begin{tabular}{@{}l@{\smallcol}r@{\smallcol}r@{\smallcol}r|r@{\smallcol}r@{\smallcol}r@{}}
\toprule
 & \multicolumn{3}{c|}{\textbf{Llama-2 7B}} & \multicolumn{3}{c}{\textbf{Llama-2 70B}} \\
\textbf{Setting names} & \textbf{EM} & \textbf{Rec.} & \textbf{R-L} & \textbf{EM} & \textbf{Rec.} & \textbf{R-L} \\
\midrule
\multicolumn{7}{c}{\emph{baselines}} \\
cont. pre-training \numone & 27.6 & 31.6 & 43.8 & 41.7 & 45.8 & 60.2\\
\quad+\ft \numtwo & 30.3 & 34.7 & 47.4 & 46.4 & 50.9 & 64.1\\
   mix all data \numfour & 39.4 & 44.6 & 56.7 & 57.1 & 63.4 & 72.4\\
\midrule
\multicolumn{7}{c}{\emph{various \preft (\pit) methods}} \\
\pit\space{\small (QA only)} \numfive & 28.6 & 32.7 & 45.2 & 49.7 & 53.7 & 67.9\\
\pit\space{\small (QA \nextphase docs)} \numsix & 32.5 & 37.2 & 49.0 & 54.6 & 60.0 & 73.8\\
\pit \numseven & \textbf{45.4} & \textbf{51.2} & \textbf{63.2} & \textbf{62.7} & \textbf{68.6} & \textbf{78.8} \\
\bottomrule
\end{tabular}
\caption{Comparison (\%) of various \preft methods versus standard \ft methods. The best results are in bold.}
\label{tab:pre}
\end{table}
\end{comment}

\begin{table*}[tb]
\small
\centering
\begin{tabular}{l@{\mediumcol}l@{\mediumcol}r@{\mediumcol}r@{\mediumcol}r}
\toprule
\textbf{Setting names} & \textbf{Setting configurations} & \textbf{EM} & \textbf{Rec.} & \textbf{R-L} \\
\midrule
\multicolumn{5}{c}{\emph{baselines}} \\
 continued pre-training \numone & test doc & 27.6 & 31.6 & 43.8 \\
\quad+\ft \numtwo & train doc + test doc \nextphase train QA & 30.3 & 34.7 & 47.4 \\
\quad+\ft (w/o forget) \numthree & train doc + test doc \nextphase train QA + test doc & 30.2 & 34.1 & 46.4 \\
\quad+\ft (w/o train doc) & test doc \nextphase train QA & 27.1 & 30.7 & 42.3 \\
weighted continued pre-training & test doc (weighted) & 27.7 & 32.7 & 43.3 \\
adapted continued pre-training & train doc \nextphase test doc & 26.9 & 32.7 & 44.2 \\
mix all data \numfour & train QA + train doc + test doc & 39.4 & 44.6 & 56.7 \\
\midrule
\multicolumn{5}{c}{\emph{various \preft (\pit) methods and ablation studies}} \\
\hlcell & \hlcell train QA + train doc (3 epochs) \nextphase test doc & \hlcell 45.4 & \hlcell 51.2 & \hlcell 63.2 \\
\hhline{~----}
\hlcell & \multicolumn{4}{c}{\hlcell \emph{ablation studies of the number of epochs}} \\
\hlcell & \hlcell \quad1 epoch & \hlcell 33.3 & \hlcell 39.1 & \hlcell 50.3 \\
\hlcell & \hlcell \quad5 epochs & \hlcell 45.8 & \hlcell 52.1 & \hlcell 63.6 \\
\hlcell & \hlcell \quad10 epochs & \hlcell 46.5 & \hlcell 52.3 & \hlcell 61.9 \\
\hhline{~----}
\hlcell & \multicolumn{4}{c}{\hlcell \emph{ablation studies of different learning mechanisms}} \\
\hlcell & \hlcell \quad QA before doc (grouped) & \hlcell 38.2 & \hlcell 43.2 & \hlcell 56.3 \\
\hlcell & \hlcell \quad QA after doc (grouped) & \hlcell 27.2 & \hlcell 31.1 & \hlcell 42.1 \\
\hlcell & \hlcell \quad QA before doc (interleaved) & \hlcell 45.9 & \hlcell 51.3 & \hlcell 64.5 \\
\multirow{-10}{*}{\hlcell \pit \numseven} & \hlcell \quad QA after doc (interleaved) & \hlcell 43.2 & \hlcell 49.1 & \hlcell 61.6 \\
\hlcelll \pit{}-{}- & \hlcelll train QA + train doc \nextphase train QA \nextphase test doc & \hlcelll 44.4 & \hlcelll 51.3 & \hlcelll 63.4 \\
\hlcellll \pit{}++ \numeight & \hlcellll train QA \nextphase train QA + train doc \nextphase test doc & \hlcellll \textbf{48.1} & \hlcellll \textbf{54.4} & \hlcellll \textbf{66.4} \\
\bottomrule
\end{tabular}
\caption{Comparison (\%) of various \preft methods and ablation studies to identify the key contributors to improved performance using Llama-2 7B. Different background colors indicate different \preft methods. The best results are in bold.}
\label{tab:ablation}
\end{table*}

\section{Improving LLMs in Absorbing Knowledge from Documents}\label{sec:pre}
The amount of knowledge elicited through the standard \ft is still limited, even though the perplexity of documents is minimized, a phenomenon we refer to as the ``perplexity curse''.
Our next question is how can we improve the ability of LLMs to absorb knowledge from documents to mitigate the perplexity curse.
The main challenge is the gap between the way knowledge is presented in raw documents and how it is accessed through question-answering.
We found that QA pairs are generally straightforward, while documents tend to be more complex and cluttered, weaving many factual statements together in a more intricate manner.
Using \autoref{fig:example} as an example, the answer to the question ``who handled the editing of Oppenheimer'' is included in a sentence in the middle of the article ``Editing was handled by Jennifer Lame ...'', which does not explicitly mention ``Oppenheimer''. 
During training, LLMs must understand the context and deduce that ``editing'' refers to ``the editing of Oppenheimer'' to effectively encode this knowledge in the parameters.

\citet{part1-zhu-2023} studied this problem by training a randomly initialized GPT-2-like transformer from scratch on synthetic biographies and evaluated its ability to answer questions about the individuals.
They found that training on a mix of biographies and questions related to half of those biographies led to strong generalization when answering questions about the remaining half of biographies, which resembles setting \numfour in \autoref{fig:settings}.
In contrast, training on biographies and QA pairs sequentially failed.
However, the key contributor to the success remains uncertain because the data were blended together, and it is unclear how to apply this practically to absorb knowledge from new documents.
Inspired by our observation of the different difficulty levels between QA pairs and documents, and the finding from \citet{part1-zhu-2023}, we hypothesize that \emph{it is beneficial to deliberately expose LLMs to \ft data before continued pre-training so that the process of encoding knowledge from complex documents takes into account how this knowledge is accessed.}
We refer to this as \textbf{\preft (\pit)} and study various implementations of \pit prior to continued learning (\autoref{sec:pre_method}), followed by detailed ablations identifying the keys contributor to performance (\autoref{sec:pre_plus} and \autoref{sec:pre_ablation}), and finally assess how well \pit performs across domains (\autoref{sec:pre_ood}).
We adhere to the hyperparameters outlined in \autoref{sec:hyper} and perform \pit for 3 epochs unless specified otherwise.

\subsection{Variants of \expandafter\MakeUppercase\preft}\label{sec:pre_method}

\paragraph{\expandafter\MakeUppercase\preft w/ QA only} We start with exposing \ft data before continued pre-training on documents---training on topically related QA pairs before training on test documents (\autoref{fig:settings} \numfive).
This can be directly compared with the continued pre-training setting (\autoref{fig:settings} \numone).
The intuition is that questions help LLMs recognize key types of information, enabling LLMs to focus on important information during pre-training on subsequent documents, even though the questions are not directly tied to the documents.
For example, training on a question like ``who handled the editing of Oppenheimer'' could help LLMs pay attention to screenwriters when training on new documents like ``Barbie''.
As shown in \autoref{tab:standard}, this method outperforms continued pre-training, especially on larger LLMs (27.6\%/41.7\% \nextphase 28.6\%/49.7\% for 7B/70B).
The ablation that trains on QA data after training on documents (``\ft w/o train doc'' in \autoref{tab:ablation}) is ineffective, confirming the importance of training on questions as a warm-up before encoding documents.

\paragraph{\expandafter\MakeUppercase\preft on QA and documents sequentially} Our second implementation trains on QA and associated documents sequentially (\autoref{fig:settings} \numsix), with the intuition that the ability to absorb knowledge from documents can be strengthened if an LLM is trained on the complex documents after it has grasped the associated simpler QA pairs.
For instance, if an LLM has already learned that ``Jennifer Lame'' is the answer to ``who handled the editing of Oppenheimer'', training on the document ``Editing was handled by Jennifer Lame'' can more efficiently refine its storage of knowledge in its parameters.
As shown in \autoref{tab:standard}, \pit on QA pairs and documents sequentially surpasses the QA-only variant (\autoref{fig:settings} \numfive) and standard \ft (\autoref{fig:settings} \numtwo) (30.3\%/46.4\% \nextphase 32.5\%/54.6\% for 7B/70B), demonstrating its effectiveness.

\paragraph{\expandafter\MakeUppercase\preft}
The effectiveness of \pit depends on ensuring that the associated QA pairs are already learned before encoding the respective documents.
However, we observed that after training on documents (train doc in \autoref{fig:settings} \numsix), the accuracy for corresponding questions (train QA in \autoref{fig:settings} \numsix) dropped from almost perfect to 30\%, indicating severe forgetting.
To fix this, we train on the associated QA pairs and documents together (\autoref{fig:settings} \numseven).
As shown in \autoref{tab:standard}, this significantly improves the performance, outperforming all other approaches, including mixing all data together (\autoref{fig:settings} \numfour), by a large margin (39.4\%/57.1\% \nextphase 45.5\%/62.7\% for 7B/70B).
Training on both QA pairs and documents prevents forgetting, but it also obscures how the learning process works.
It is unclear whether LLMs grasp QA pairs before encoding knowledge from documents, or if it works the other way around.
In the following section, we deliberately arrange the order of QA pairs and documents during training to examine this, which leads us to propose an improved version of \pit.

\subsection{\expandafter\MakeUppercase\preft{}++}\label{sec:pre_plus}
We first study how the performance varies with different numbers of epochs.
As shown in \autoref{tab:ablation}, training for 1 epoch is insufficient, and the performance of 3, 5, or 10 epochs is similar.
We fix the number of epochs to 3 and arrange the order of QA pairs and corresponding documents as shown in \autoref{fig:order} in \autoref{sec:app_ablation}.
The interleaved arrangement cycles through all the data 3 times, ensuring that in each epoch, questions either precede or follow their associated documents.
On the other hand, the grouped arrangement clusters each example's 3 appearances together, guaranteeing that the repeated questions are positioned either before or after their respective repeated documents.
As shown in \autoref{tab:ablation}, positioning QA pairs before corresponding documents achieves better performance in both grouped and interleaved arrangements, indicating that during \pit, the learning mechanism prioritizes understanding how to access knowledge before learning to absorb information from the more complex and information-dense documents.

Based on this, we propose an improved variant called \preft{}++, which trains exclusively on QA pairs to understand patterns of knowledge access, then progresses to training on a combination of QA and document data to align knowledge access through questions and knowledge encoding from documents (\autoref{fig:settings} \numeight).
As shown in \autoref{tab:ablation}, \pit{}++ significantly outperforms \pit (\autoref{fig:settings} \numseven) from 45.4\% to 48.1\%, while training on QA data after on the mix (\pit{}-{}- in \autoref{tab:ablation}) does not yield additional benefits.
This reinforces our hypothesis that understanding how knowledge is accessed aids in absorbing knowledge from documents, and therefore, should be prioritized.

\subsection{Ablation Studies}\label{sec:pre_ablation}
\paragraph{Standard \ft is inferior not due to forgetting} A drawback of standard \ft is that knowledge in test documents might be forgotten after training on QA pairs (a phenomenon also known as the ``alignment tax'' \cite{instructgpt3-ouyang-2022}).
To show that the lower performance of standard \ft is not due to forgetting, we add a setting where we mix train QA with test documents during \ft to prevent forgetting (\autoref{fig:settings} \numthree).
As shown in \autoref{tab:ablation}, this does not help, confirming our hypothesis.

\paragraph{\expandafter\MakeUppercase\preft is not simply upweighting salient tokens from documents} We include an ablation inspired by \citet{camels-hu-2023} which upweights tokens when pre-training on documents to focus on salient information.
We assign a weight of 1.0 to tokens in documents that are included in the answers (e.g., ``Jennifer Lame'' in the sentence ``Editing was handled by Jennifer Lame''), and assign a lower weight of 0.5 to other tokens.
As shown in \autoref{tab:ablation}, this weighted continued pre-training is ineffective, confirming our hypothesis.

\begin{table}[tb]
\small
\centering
\begin{tabular}{lr@{\mediumcol}r@{\mediumcol}r|r@{\mediumcol}r@{\mediumcol}r}
\toprule
 & \multicolumn{3}{c|}{\textbf{Llama-2 7B}} & \multicolumn{3}{c}{\textbf{Llama-2 70B}} \\
\textbf{Settings} & \textbf{EM} & \textbf{Rec.} & \textbf{R-L} & \textbf{EM} & \textbf{Rec.} & \textbf{R-L} \\
\midrule
\multicolumn{7}{c}{\emph{standard \ft} \numtwo} \\
in-domain & 30.3 & 34.7 & 47.4 & 46.4 & 50.9 & 64.1 \\
cross-domain & 23.6 & 28.2 & 38.4 & 42.8 & 49.7 & 58.5 \\
\midrule
\multicolumn{7}{c}{\emph{\preft} \numseven} \\
in-domain & 45.4 & 51.2 & 63.2 & 62.7 & 68.6 & 78.8 \\
cross-domain & 36.9 & 43.2 & 54.9 & 55.2 & 66.7 & 74.0 \\
\bottomrule
\end{tabular}
\caption{In-domain and cross-domain \pit.}
\label{tab:ood}
\end{table}

\begin{table}[tb]
\small
\centering
\begin{tabular}{lrrr}
\toprule
\textbf{Settings} & \textbf{EM} & \textbf{Rec.} & \textbf{R-L} \\
\midrule
\multicolumn{4}{c}{\emph{generalization to the biography dataset \dataset{bioS} }} \\
closed-book & 2.9 & 2.9 & 11.0 \\
open-book w/ doc & 95.2 & 95.4 & 95.6 \\
continued pre-training \numone & 29.6 & 29.8 & 38.7 \\
\preft \numseven & \textbf{58.1} & \textbf{58.4} & \textbf{61.9} \\
\midrule
\multicolumn{4}{c}{\emph{generalization to questions by real users from Google }} \\
standard \ft \numtwo & 21.5 & 30.1 & 36.8 \\
\preft \numseven & \textbf{29.0} & \textbf{35.5} & \textbf{48.2} \\
\bottomrule
\end{tabular}
\caption{Generalization of the Llama-2 7B model trained with \preft.}
\label{tab:real}
\end{table}

\subsection{Cross-domain Generalization}\label{sec:pre_ood}
We validated the effectiveness of \pit by training and evaluation on data from the same domain (\dataset{Wiki2023-film}).
\emph{Can \pit make LLMs better at absorbing knowledge from documents of a different domain?}
To this end, we follow the cross-domain setting outlined in \autoref{fig:dataset}---training on other domains (\dataset{Wiki2023-other-train}) and testing on the film domain (\dataset{Wiki2023-film-test}).
The results of standard \ft and \pit, in both in-domain and cross-domain settings, are detailed in \autoref{tab:ood}.
Even though it is not as effective as the in-domain counterparts, cross-domain \pit still significantly outperforms \ft, demonstrating that it can generalize across different domains.
This finding sheds light on the potential to scale this method up to a broader range of documents and instructions for more robust generalization.

We also evaluate the effectiveness of \pit in two other scenarios: (1) when applied to non-Wikipedia documents, and (2) when addressing questions asked by real users.
For the first scenario, we take the Llama-2 7B model trained with \pit on \dataset{2023Wiki-other} and further train it on biographies synthesized in \citet{part1-zhu-2023} (\dataset{bioS}).
Then, we evaluate based on questions about the individuals.
For the second scenario, we manually search Google using questions generated by LLMs from \dataset{Wiki2023-film-test}, collect a total of 93 similar questions from real users by leveraging Google's ``People Also Ask'' feature, and then evaluate Llama-2 7B on these questions.
As shown in \autoref{tab:real}, \pit outperforms baselines in both scenarios, demonstrating its generalization ability.

\section{Related Work}

\subsection{Continual Knowledge Acquisition}
Several works have studied whether LMs can answer questions about information in documents they have been trained on.
\citet{lmkb-wang-2021,continual-jang-2023,camels-hu-2023} use relatively small LMs such as BART \cite{bart-2020-lewis}, T5 \cite{t5-raffel-2020}, or GPT-2 \cite{radford-2019-gpt2}.
\citet{ftrag-ovadia-2023} focus on the comparison between RAG and continued pre-training approaches without using \ft.
\citet{part1-zhu-2023,part2-zhu-2023} examine this problem from a similar angle as ours using a GPT-2-like transformer trained from scratch on synthetic biographies and fine-tuned on QA pairs related to the individuals.
They examined a mixed training setting on both biographies and QA pairs, which is our major motivation to study different strategies to incorporate QA data before continued pre-training.
Other works study adapting LLMs to new domains via various strategies \cite{domainchat-2023-zhang,readadapt-2023-cheng,medalpaca-2023-han,pmcllama-2023-wu,astrollama-2023-nguyen,zhao-llm-2023}.

\subsection{Instruction-tuning or Alignment}
\expandafter\MakeUppercase\ft (also known as supervised fine-tuning) on high-quality annotated data \cite{t0-sanh-2022,flan-jason-2022,crosstask-2022-mishra,iyer-optiml-2022,openassistant-kopf-2023,lima-zhou-2023,principle-zhiqing-2023,salmon-2023-sun} and/or data generated by proprietary models \cite{alpaca-taori-2023,vicuna-2023,tulu-2023-wang,camel-2023-ivison}, or alignment with reinforcement learning from human feedback (RLHF) or direct preference optimization (DPO) \cite{instructgpt3-ouyang-2022,llama2-touvron-2023,dpo-rafailov-2023,factft-tian-2023} has been a central topic recently because it elicits knowledge from LLMs and enhances various abilities to handle questions from users.
We focus on factuality and study the best way to perform \ft to elicit factual knowledge from LLMs.

\subsection{Analyzing the Training Dynamics of LMs}
Many works study the training dynamics of LMs from different perspectives.
\citet{quantifylmmemory-carlini-2022} quantifies memorization across model sizes and the frequency of data duplication.
\citet{memorization-tirumala-2022} finds that larger LMs memorize training data faster with less overfitting.
\citet{trajectory-xia-2023} shows that perplexity is more predictive of model behaviors than other factors.
\citet{meat-lucio-2022} studies end-task aware pre-training using classification tasks and RoBERTa models.
\citet{qapretrain-jia-2022} adds a pre-training objective to encourage the vector for each phrase to have high similarity with the vectors for all questions it answers.
Our work differs in that we specifically focus on the capacity of recalling and generalizing information from a seen document to answer questions.

\subsection{Retrieval-augmented Generation}
Retrieval-augmented generation (RAG) is a widely used approach to incorporate new knowledge into LLMs by augmenting fixed LLMs with retrieved information from external sources \cite{drqa-chen-2017,realm-guu-2020,rag-lewis-2020,retro-borgeaud-2022,retro-wang-2023,retomaton-alon-2022,effknnlm-he-2021,emdr2-2021-sachan,atlas-izacard-2022,yono-2021-lee,reatt-jiang-2022,replug-shi-2023,flare-jiang-2023,selfrag-asai-2023,webgpt-nakano-2021,webcpm-qin-2023,radit-lin-2023}.
While RAG is effective in reducing hallucinations commonly experienced when relying solely on knowledge stored in parameters, its retrieval and generation process adds extra latency and complexity.
In contrast, continued pre-training to store knowledge in parameters and utilizing the stored knowledge to answer questions in a closed-book manner are simpler and faster at inference time.
Enhancing this capability is also scientifically significant, as it represents a fundamental step in employing LLMs as dependable assistants for accessing information.
Therefore, this paper focuses on exploring parametric approaches.

\section{Conclusion}
We study the best way of continued training on new documents with the goal of later eliciting factual knowledge and propose \preft that learns how knowledge is accessed via QA pairs prior to encoding knowledge from documents.
Extensive experiments and ablation studies demonstrate the superiority of \preft versus standard \ft.
Future directions include scaling this method up to a broader range of documents and instructions for more robust generalization.

\section{Limitations}
The \dataset{Wiki2023} dataset provides a relatively clean testbed for studying continual knowledge acquisition.
However, its scope is limited to Wikipedia, which restricts the trained models' adaptability to other sources like web pages from Common Crawl or scientific documents from arXiv.
We focus on eliciting factual knowledge with \ft on QA data in this paper.
The effectiveness of \preft with different types of data for enhancing other skills like reasoning or comprehension is something that needs to be explored in future studies.

\section*{Acknowledgements}
We would like to thank Zeyuan Allen-Zhu, Zexuan Zhong, Shuyan Zhou, Frank F. Xu, Qian Liu, and Ruohong Zhang for their help with the experiments and constructive feedback.

\bibliography{custom}
%\newpage

\appendix
\section{\dataset{Wiki2023} Dataset}\label{sec:app_wiki2023}
\begin{prompt}[title={Prompt \thetcbcounter: question-answer generation prompt}, label=prompt:qa]
Given the following summary about the subject \{topic\}, generate a comprehensive list of questions and corresponding answers that cover all aspects. To make the question clear, always include \{topic\} in the question. Answers should be concise, consisting of a few short phrases separated by commas.\\
\\
Output in the following format:\\
Q: an open-domain question about the subject \{topic\} (the subject \{topic\} should always be included)\\
A: phrase1, phrase2, ...\\

Summary:\\
\{summary\}
\end{prompt}

\section{Hyperparameters}\label{sec:app_hyper}
We use AdamW \cite{adamw-loshchilov-2019} with $\beta_1=0.9$, $\beta_2=0.95$, and a weight decay of 0.1.
We decay the learning rate to 10\% of its initial value using a cosine scheduler without warm-up.
When pre-training on documents, we use a batch size of 256 documents and an initial learning rate of 3e-5.
During \ft on QA pairs, we use the same batch size of 256 QA pairs, but opt for a reduced initial learning rate of 5e-6 because the number of tokens in a single batch used for computing losses is lower.
The number of epochs varies depending on the setting and is detailed in the corresponding sections.

\section{Evaluation Metrics}\label{sec:app_metric}
At inference time, we use greedy decoding to generate answers given questions as context, following the format in \autoref{fig:example}.
To evaluate the original Llama-2, we add 5 QA pairs as in-context exemplars to make sure it follows the QA format.
Since most questions are simple factoid questions and most answers are relatively short, we use exact match (EM) as our primary metric \cite{nq-kwiatkowski-2019}, which measures whether the model's output matches the gold answer exactly after normalization (e.g., remove articles and punctuations).
To assess longer responses and accommodate minor lexical differences, we also report answer recall, which assesses if the gold answer appears in the model's output, and ROUGE-L, which measures the longest common subsequence between the model's output and the gold answer.

\begin{figure}[tb]
\includegraphics[width=1.0\columnwidth, clip, keepaspectratio]{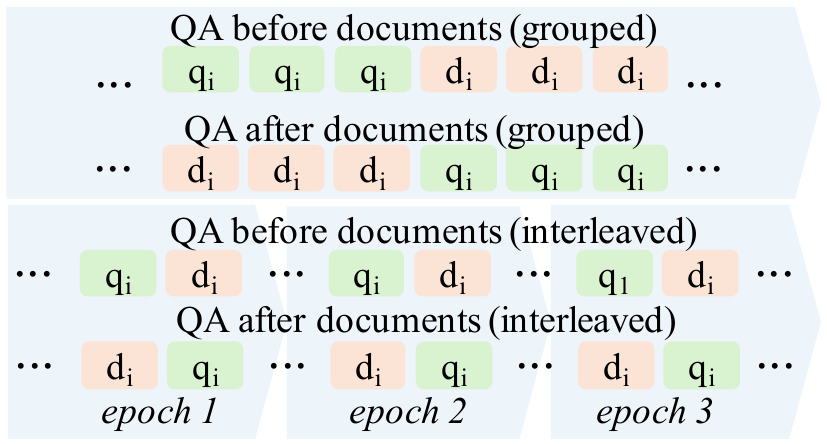}
\centering
\caption{Different arrangements between QA pairs and corresponding documents. The ellipses represent other examples.}
\label{fig:order}
\end{figure}

\section{Details of Ablation Studies}\label{sec:app_ablation}
We arrange the order of QA pairs and corresponding documents as shown in \autoref{fig:order} to study the learning mechanism of \preft. 

\end{document}